\documentclass[conference]{IEEEtran}
\usepackage{bm}
\usepackage{listings}
\usepackage{multirow}
\usepackage{pifont}
\usepackage{xcolor}
\usepackage{hyperref}
\usepackage{graphicx}
\usepackage{subcaption}
\usepackage{graphicx}
\usepackage[most]{tcolorbox}

\usepackage{tikz}
\usetikzlibrary{babel}
\usepackage{enumitem}

\newtcolorbox{enumhighlight}{
  colframe=black,
  colback=white,
  boxrule=0.8pt,
  arc=2pt,
  left=2pt,
  right=6pt,
  top=4pt,
  bottom=4pt,
  before skip=0pt,
  after skip=0pt
}

 \DeclareRobustCommand*\circled[1]{%
  \tikz[baseline=(char.base)]{%
    \node[
      circle,
      draw,
      line width=0.9pt,
      inner sep=0pt,
      minimum size=2.2ex,
      align=center
    ] (char) {\footnotesize\bfseries #1};
  }%
 }

\newtcolorbox{promptbox}{%
	floatplacement=h,
	colback=gray!10,
	colframe=black,
	arc=2pt,
	boxrule=0.7pt,
	left=8pt,
	right=8pt,
	top=8pt,
	bottom=8pt,
 }

\newcommand{\stitle}[1]{\bigskip\noindent\textbf{#1}}

\def\bikmode{1}
\newcommand{\bik}[2][BK]{%
  \ifnum\bikmode=1
    \textcolor{orange}{{[#1:}} {\color{orange}{#2}}\textcolor{orange}{{]}}%
  \fi
}

\def\dsmode{1}
\newcommand{\ds}[2][DS]{%
  \ifnum\dsmode=1
    \textcolor{purple}{\textbf{[#1:}} {\color{purple}{#2}}\textcolor{purple}{\textbf{]}}%
  \fi
}

\begin{document}

\hypersetup{
	colorlinks,%
	citecolor=black,%
	filecolor=black,%
	linkcolor=black,%
	urlcolor=black,%
	pdfborder={0 0 0}, colorlinks=false, allcolors=black,
	pdftitle={Context-Enriched Natural Language Descriptions of Vessel Trajectories},
	pdfauthor={Kostas Patroumpas, Alexandros Troupiotis-Kapeliaris, Giannis Spiliopoulos, Panagiotis Betchavas, Dimitrios Skoutas, Dimitris Zissis, Nikos Bikakis},
	pdfsubject={Maritime AI, Vessel Trajectory Analysis, Large Language Models, Semantic Trajectories, SpatioTemporal Data Mining, Explainable AI, Trajectory Annotation, Spatial Data Enrichment, Mobility Pattern Mining},
	pdfkeywords={Maritime AI, Vessel Trajectory Analysis, Large Language Models, Semantic Trajectories, SpatioTemporal Data Mining, Explainable AI, Trajectory Annotation, Spatial Data Enrichment, Mobility Pattern Mining, Transportation Foundation Models, Mobility Intelligence, Automatic Identification System}
}

\title{Context-Enriched Natural Language Descriptions \\ of Vessel Trajectories}

\author{\IEEEauthorblockN{
Kostas Patroumpas\IEEEauthorrefmark{1} \hspace{1em}
Alexandros Troupiotis-Kapeliaris\IEEEauthorrefmark{2}\IEEEauthorrefmark{1} \hspace{1em}
Giannis	Spiliopoulos\IEEEauthorrefmark{2}\IEEEauthorrefmark{1} \\
Panagiotis Betchavas\IEEEauthorrefmark{1}   \hspace{0.5em}
Dimitrios Skoutas\IEEEauthorrefmark{3}\hspace{0.5em}
Dimitris Zissis\IEEEauthorrefmark{2} \hspace{0.5em}
Nikos Bikakis\IEEEauthorrefmark{4}\IEEEauthorrefmark{1}
}
\IEEEauthorblockA{
\IEEEauthorrefmark{1}Archimedes/Athena RC, Greece
\IEEEauthorrefmark{2}University of the Aegean, Greece
\IEEEauthorrefmark{3}Athena RC, Greece}
\IEEEauthorblockA{
\IEEEauthorrefmark{4}Hellenic Mediterranean University, Greece}
}

\maketitle

\IEEEpeerreviewmaketitle

\begin{abstract}
We address the problem of transforming raw vessel trajectory data collected from AIS into structured and semantically enriched representations interpretable by humans and directly usable by machine reasoning systems. We propose a context-aware trajectory abstraction framework that segments noisy AIS sequences into distinct trips each consisting of clean, mobility-annotated episodes. Each episode is further enriched with multi-source contextual information, such as nearby geographic entities, offshore navigation features, and weather conditions. Crucially, such representations can support generation of controlled natural language descriptions using LLMs. We empirically examine the quality of such descriptions generated using several LLMs over AIS data along with open contextual features. By increasing semantic density and reducing spatiotemporal complexity, this abstraction can facilitate downstream analytics and enable integration with LLMs for higher-level maritime reasoning tasks.

\end{abstract}


\maketitle


\section{Introduction}
\label{sec:introduction}


Modern sensor and mobile devices have enabled an unprecedented surge in availability and accumulated volume of trajectory data collected for various applications, calling for more advanced methods for their efficient management and analysis. 
In particular, tracking vessels has been greatly facilitated by the \emph{Automatic Identification System} (\emph{AIS}), which provides their identity, position, course, heading, etc. in real time, and thus can support crucial maritime tasks such as route
planning, course prediction, or anomaly detection \cite{yang2019big, yang2024harnessing}.
While AIS messages offer precise location information, they are inherently noisy, occasionally sparse, and crucially lack explicit semantic annotation. Since raw AIS trajectories are sequences of spatiotemporal tuples devoid of contextual grounding, relational abstractions, mobility event characterization, and domain-specific knowledge, direct interaction between unprocessed AIS data with modern \emph{Large Language Models} (LLM) is severely hindered~\cite{xu2025trajectorypredictionmeetslarge}.




To address this limitation, we propose a methodology that segments AIS trajectories into distinct trips and then augments them with contextual knowledge derived from heterogeneous data sources. Our approach adheres to the paradigm introduced in ~\cite{10.1145/2483669.2483682,10.1145/2501654.2501656} for constructing \emph{semantic trajectories}, but applied specifically for vessels. Each trip represents a sequence of \textit{episodes} that abstract vessel mobility patterns (e.g.,  stopped, turning, sailing). Regarding \emph{context}, geographic layers (e.g., protected areas, capes, straits), data specific to the maritime domain (such as ports, anchorage zones, shipping lanes, traffic separation schemes, and bathymetry), as well as weather forecasts (e.g., wind force and direction) can be utilized. Thus, vessel movements get enriched with semantic \emph{annotations} describing important interactions (e.g., port calls, strait transits, proximity to environmentally sensitive zones). By explicitly linking trajectory segments to contextual features and inferred mobility events, our framework enhances interpretability and explainability, enabling transparent reasoning about why a vessel followed a particular course or exhibited irregular behavior (e.g., a detour).

Most importantly, building upon such enriched trajectory representations, we are able to generate human- and machine-readable \emph{natural language narratives} of vessel behavior 
using LLMs. Semantic trajectories embed prior knowledge about navigational corridors, port operations, sea depth limitations, and traffic regulatory zones, thereby providing structured inductive biases. In effect, input sequences of raw AIS messages are transformed into context-aware movement narratives that distill mobility features and better reflect navigational intent, operational patterns, and regulatory guidelines. 
Our empirical investigation with several LLMs indicates that using such contextually-enhanced trajectory descriptions can improve robustness, reduce uncertainty in case of sparse or noisy AIS data, and enable a more principled detection of deviations from expected maritime behavior. As such, semantic enrichment serves not only as an interface for LLM-based reasoning but also as a foundation for more accurate and explainable maritime analytics. This ensures compatibility with automated processing pipelines while preserving accessibility for maritime stakeholders such as port authorities, coast guards, or shipping companies. 



The generated descriptions can have several applications in the maritime domain. 
They may assist in contextualizing possible events, such as behavior anomalies (e.g., low speed, abrupt turning) or accidents (e.g., groundings) \cite{bakdi2019ais}.
Interestingly, such descriptions resemble \emph{sailing route directions} consisting of safe passage guidelines or alerts, coastal or port approach suggestions, etc., based on coordinates of known landmarks (e.g., lights, buoys).
For instance, detailed suggestions are stipulated for the passage shown in Figure~\ref{fig:trip-annotated-example} in Baltic Sea\footnote{\href{https://msi.nga.mil/api/publications/download?key=16694491/SFH00000/Pub194bk.pdf&type=view}{\scriptsize https://msi.nga.mil/api/publications/download?key=16694491/SFH00000/Pub194bk.pdf}}. 
Comparing our proposed context-enriched narratives with such official suggestions can be a good indicator of possible deviations throughout a trip, and also an important step towards maintaining interoperability of suggestions crucial for maritime safety.
In addition, such descriptions can automatically provide \emph{voyage reports}\footnote{\href{https://support.marinetraffic.com/en/articles/9552765-voyage-report}{\scriptsize https://support.marinetraffic.com/en/articles/9552765-voyage-report}}, allowing shipping companies to inspect each leg of a trip, check whether the itinerary of a vessel was compliant to charter agreements, and maintain historical logs for route planning in future trips.
Last, but not least, textualized semantic trajectories can enable LLMs to reliably ingest, reason over, and synthesize maritime movement information for \emph{downstream analytical tasks}. For example, more accurate trajectory prediction~\cite{chen2025semint} can be achieved by taking advantage of context to infer voyage intentions. Our context-enriched narratives ingest extra knowledge that may  improve imputation of missing trajectory segments~\cite{liu2026vista,habitedbt}. Textual descriptions can also assist in navigation and collision avoidance tasks~\cite{ma2025navigation} that require complex reasoning to make accurate decisions.


Several recent frameworks have employed LLMs to extract~\cite{park2025ais,chen2025semint} or process~\cite{you2025interpretable} vessel trajectory narratives. However, their scope is mostly on maritime tasks like making collision avoidance decisions through textualized navigation scenarios~\cite{ma2025navigation}, simply inferencing origin-destination ports~\cite{chen2025semint}, or apply a cluster-based trajectory segmentation to create semantic descriptions~\cite{liu2025vtllm}. 
Instead, our approach detects vessel mobility events to construct semantic trajectories, annotates them with rich contextual knowledge from geographic, maritime, and weather data sources, and ultimately uses LLMs to distill them into reliable, concise, and factual trip narratives.



Our contributions can be summarized as follows:

\begin{itemize}
    \item We develop a framework for constructing semantic trajectory representations in episodes
    that summarize mobility events along a vessel's course from raw AIS locations.

    \item We further enrich each semantic episode with context from various geographic, maritime, and meteorological data sources in various formats to provide deep insights into the motion patterns of vessels.

    \item We leverage LLMs to accept such semantic representations and generate textual descriptions of vessel trips that are both human- and machine-readable. 
    
    \item We conduct a  empirical study over real-world AIS data enriched with context from various open data sources to generate trajectory descriptions in natural language using LLMs of varying capabilities. 

    \item We evaluate the quality of such descriptions in terms of relevance, faithfulness, and correctness, providing strong evidence that such AI-generated narratives can help stakeholders explain and interpret vessel behavior at sea. 
    
\end{itemize}

The remainder of this paper proceeds as follows. Section~\ref{sec:related} surveys related work. In Section~\ref{sec:methodology} we present our framework for constructing contextually-enriched semantic trajectories of vessels. Section~\ref{sec:generator} explains how LLMs can then be used to generate textual descriptions from such trajectory representations. Section~\ref{sec:evaluation} reports results from our empirical evaluation. Section~\ref{sec:conclusion} offers conclusions and future research directions.

\section{Related Work}
\label{sec:related}

Trajectory data management offers methods for cleaning, storage, indexing, mobility analytics and mining, and certainly support for diverse types of queries (e.g., spatial-only, spatio-temporal,  spatio-textual) over trajectories. Recent surveys~\cite{10.1145/3440207, atluri2018spatio} offer a comprehensive review of relevant research trends.


Recent advances in ML models designed for Natural Language Processing have also attracted research interest in the mobility domain. Typically, discretized location identifiers act as tokens (like words) and full trajectories as sequences (like sentences)~\cite{musleh2024let}. Architectures based on BERT \cite{devlin2019bert} have been adapted for such tokenized trajectory sequences and develop specialized models like TrajBERT~\cite{si2023trajbert}, Kamel~\cite{musleh2023kamel} or BERT4Traj~\cite{yang2025bert4traj}. Yet, such models necessitate domain-specific vocabularies, which 
neglects the deep relational knowledge in emerging LLMs.
Indeed, leveraging LLMs for direct transformation of trajectories into textual entries can take advantage of their sophisticated reasoning and contextual depth to interpret intricate relationships in input data~\cite{liang2025foundation}. Prompt engineering is crucial for successfully translating the mobility task and input data into semantic descriptions~\cite{zhang2024large} in order to effectively answer queries like next location prediction or history summarization~\cite{wang2023would}.

Specifically in the maritime domain, several models have recently been proposed. For collision risk assessment, \cite{park2025ais} leverages a multi-scale time-series encoder, a maritime-specific text-based encoder, along with a cross-modality alignment module in order to provide numerical forecasts along with explanations in natural language. By integrating multiple heterogeneous data sources (e.g., AIS, remote sensing imagery, structured metadata) for improved maritime situational awareness, the transformer-based multimodal model proposed in \cite{wang2026towards} performs vessel detection and interpretable navigation compliance reasoning. Through trajectory segmentation into  episodes, \cite{guo2025natural} creates semantic tokens, and then uses an LLM-based pipeline to interpret natural language queries and respond through semantic matching. Without generating textual trajectory descriptions but simply via natural language interfaces, \cite{merten2025using} examines the reasoning capabilities of LLMs in querying maritime datasets. However, generating comprehensive trip descriptions can significantly enhance the contextual awareness of LLMs. As studied in~\cite{liu2025vtllm}, utilizing full narratives that integrate nearby Points-of-Interest, LLMs can be adapted to predict future movements within a rich geographic context. SEMINT~\cite{chen2025semint} leverages natural language narratives to infer the vessel’s navigation intent, enabling more accurate and context-aware long-term trajectory predictions. Towards offering collision avoidance recommendations, Navigation-GPT~\cite{ma2025navigation} encodes real-time vessel states into the language space to align encounter scenarios with maritime regulations. Textual descriptions can also assist in interpreting vessel intent~\cite{you2025interpretable} and detecting anomalous behavior \cite{mbuya2024trajectory}, as demonstrated in MSCE \cite{chen2025msce}, which utilizes multi-modal semantic cognition to uncover trajectory irregularities.

In this work, we propose a framework that generates textual narratives of vessel trajectories using LLMs. Our work applies approaches for mobility data segmentation and annotation towards construction of \emph{semantic trajectories}~\cite{10.1145/2483669.2483682,10.1145/2501654.2501656} in order to detect motion patterns and enable movement interpretation.
According to this generic paradigm, spatio-temporal features are converted into structured trajectories (as stop and move episodes), also utilizing application-dependent geographic context (e.g., road network, Points of Interest) to annotate stops and moves with the relevant objects.
In our case, from raw AIS data we identify pivotal mobility episodes and enrich them with contextual knowledge from heterogeneous data sources (geographic, maritime, meteorological). 
Ultimately, this distilled information can serve as a foundational layer for broader maritime AI architectures, providing such standardized semantic descriptors as input to task-specific models for vessel trajectory analysis.


\section{Constructing Context-Enriched Semantic Trajectory Representations}
\label{sec:methodology}

In this section, we present our open-source framework\footnote{\href{https://github.com/M3-Archimedes/AIS-semantic-trajectories}{https://github.com/M3-Archimedes/AIS-semantic-trajectories}} that accepts as input raw AIS locations and constructs context-enriched trajectory representations. It first annotates important points along the course of each vessel according to detected mobility events (e.g., stops, turns, slow motion). Based on such annotations, it then segments each trajectory into separate trips each consisting of a sequence of episodes. Finally, it ingests extra context from various (geographical, maritime, meteorological) sources into each episode. The resulting semantic trajectory representation can be exported into several formats 
for further processing, analytics, or visualization in maps and charts. Note that no LLM is used in any stage of this workflow.


\subsection{Trajectory Annotation with Mobility Events}
\label{sec:annotation}


In order to identify critical locations along a vessel's trajectory, our method  employs the \emph{trajectory compression} framework proposed in~\cite{10.1007/s10707-022-00475-0}, which is specifically tailored for annotating AIS data. This maintains a velocity vector over the recent history of each vessel's movement and applies spatiotemporal filters and rules strictly over raw AIS messages by considering factors such as time, location, as well as changes in speed and heading. By examining how the motion pattern of a given vessel changes across time, this method can semantically annotate selected positions that signify several types of \emph{mobility events}:

\begin{itemize}
\item \emph{Stops} indicate time periods where the vessel remains stationary, i.e., its speed $v<0.5$ knots. Although stops mostly occur in port areas, they can be occasionally observed offshore or even at open sea (e.g., due to mechanical failures or suspicious activities).

\item \emph{Gaps} in communication indicate that no AIS signal has been reported from a vessel over some time,
e.g., in the past $\Delta T = 30$ minutes. 

\item {\em Speed change} indicates that the vessel has been decelerating or accelerating. If the rate of speed change exceeds a given threshold $a$ (e.g., $a$ = 25\%), this marks a significant change in speed, typically occurring when the vessel reaches or departs from a port or makes a maneuver.

\item {\em Slow motion} means that the vessel consistently moves at low speed (e.g., $<$ 1 knot) over some time. In contrast to a stop event, these successive locations usually occur along a path while the vessel is sailing.

\item \emph{Turns} are detected once the actual heading over ground deviates significantly, e.g., above a threshold $\theta > 5^{o}$. This comparison is done with respect both to the previous heading (for abrupt changes) but also with the velocity vector (for smooth, long-lasting turns typical for large ships). In case the vessel is stopped or moves slowly, any changes in heading can be ignored as they are likely caused by sea drift.
\end{itemize}

\begin{figure}[t]
    \centering
\includegraphics[width=0.95\linewidth]{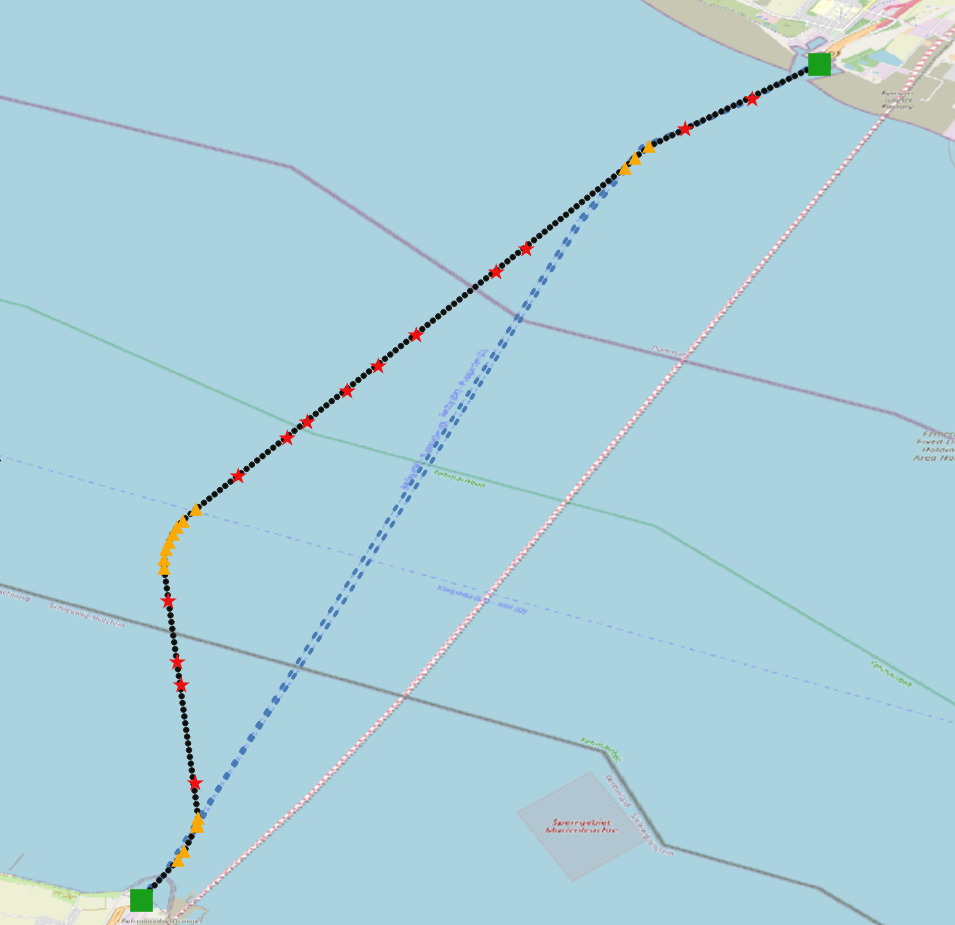}
    \caption{Originally reported AIS locations (in black dots) of a vessel after annotation: \textcolor{green}{green} boxes signify stops, \textcolor{orange}{orange} triangles denote turning points, and \textcolor{red}{red} stars indicate noise.}
    \label{fig:trip-annotated-example}
\end{figure}

In our processing workflow, we have significantly improved this method\footnote{\href{https://github.com/M3-Archimedes/AIS-trajectory-annotation}{https://github.com/M3-Archimedes/AIS-trajectory-annotation}}; instead of compression, it is now specifically geared towards more efficient, rich annotations of pivotal locations along each vessel's course. Crucially, we can apply parametrizations according to ship types, e.g., specifying different angle threshold for smaller crafts and large cargo or passenger vessels. In addition, for more accuracy in spatiotemporal calculations, the velocity vector maintained per vessel can now be defined over its last few locations (e.g., 10 previous AIS messages) or over a recent time-based sliding window (e.g., 10 minutes back from now). Furthermore, instead of a single characterization, a given location can carry multiple annotations (e.g., both  change in heading and slow motion), which is important in identifying successive episodes along a trajectory. Finally, this process also applies suitable filters that eliminate much of the noise inherent in AIS messages, such as duplicate positions, invalid coordinates, delayed messages, etc. Figure~\ref{fig:trip-annotated-example} illustrates annotated AIS locations along the trajectory of a passenger vessel that makes an itinerary between two ports in Denmark. Note that each turn includes multiple successive points, typical for large ships like this one.


\begin{figure*}[!t]
\centering 
\includegraphics[width= \textwidth]{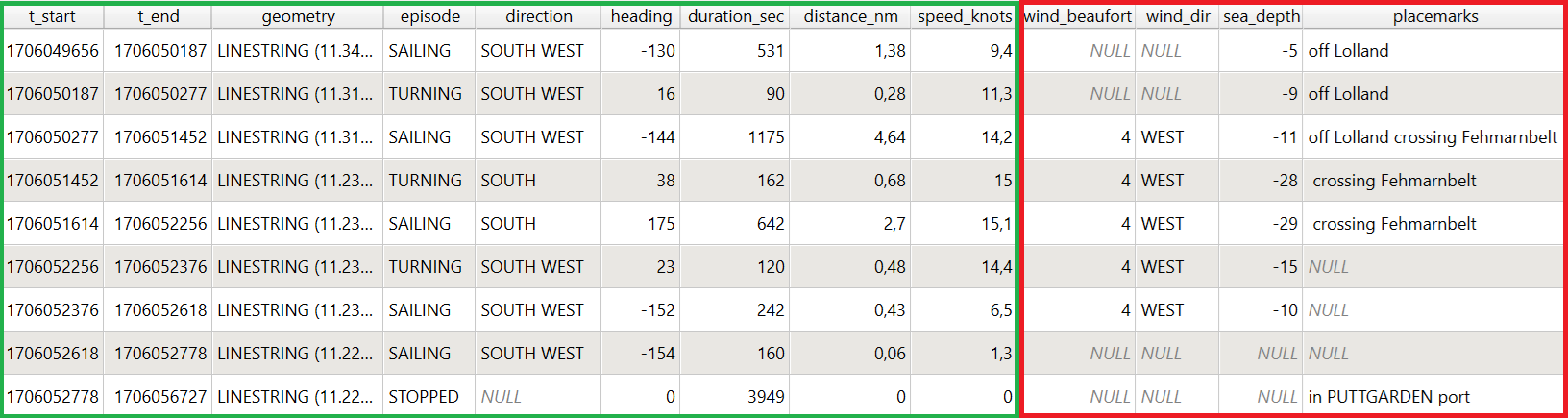}
\caption{Trip representation in successive episodes with statistics (in green box) along with contextual information from various external data sources (in red box) as extracted in CSV format suitable for map visualization (MAP).} 
\label{fig:example-trip-csv}
\end{figure*}

\subsection{Trajectory Segmentation in Trips and Episodes}
\label{sec:segmentation}

The accumulated sequence of AIS locations per vessel can eventually become very long over time, given that a ship may relay even thousands of messages per day. Splitting a trajectory into smaller parts can greatly facilitate its indexing and subsequent processing in large maritime repositories. We propose segmenting the trajectory of a vessel into distinct \emph{trips}, each one consisting of successive \emph{episodes}, by taking advantage of the aforementioned annotations by mobility events. 


\stitle{Trips.}
In our workflow, a \textit{trip} comprises the subsequence of AIS locations between two successive stops or long gaps. More specifically, the segmentation process  iterates chronologically over the annotated locations of a given vessel. Noisy locations are discarded, and each annotated location concerning stop or gap events is examined as follows:

\begin{itemize}
    \item The last location in a stop event signifies that the vessel will next be departing for a new trip.
    \item The first location in a stop event  marks the end of the current trip.
    \item If a long gap occurs (e.g., lasting more than a given $\delta =$ 3 hours), the current trip is abruptly terminated. 
    \item After such a long gap, once communication with the vessel is resumed, a new trip is assigned.
\end{itemize}

In effect, locations of a vessel between two successive stops or long gaps are grouped together with a unique trip identifier. Thus, trip origin and destination are not defined strictly by stops (typically in a port, but sometimes at open sea), but also by long-lasting gaps. 
Note that in trip segmentation we set a longer gap duration $\delta > \Delta T$ used in annotating a gap event (as discussed in Section~\ref{sec:annotation}). This is a deliberate decision (subject to a user-specified parameter $\delta$) to account for cases where contact with a vessel is temporarily lost in the middle of its voyage, e.g., at open sea with no AIS coverage, yet it effectively continues the course to its destination. Thus, smaller communication gaps lasting less than $\delta$ may still occur in the designated trips. 

Several \emph{statistics} can be calculated per trip, such as its total \emph{duration}, \emph{travelled distance}, and \emph{average speed}, accurately based on the original AIS reports.

\stitle{Episodes.}
In principle, an \textit{episode} abstracts a subsequence of AIS locations that are collectively characterized with a particular mobility event (i.e., a motion pattern like stop, turn, or gap), because they are highly correlated with respect to spatio-temporal feature (e.g., speed, heading, time period).
While iterating over the annotated locations per vessel, distinct episodes within a trip can be identified: 

\begin{itemize}
    \item \emph{STOPPED}: This  spans an entire stop event; the centroid of those locations is used to represent the episode.
    \item \emph{TURNING}: This includes a series of successive locations: it starts from a location annotated with a significant change in heading and ends up to the last similarly annotated point before the vessel course stabilizes to a new fixed heading (i.e., after the turn). Isolated turning points are discarded as noise as they are extremely rare for most vessels, which typically take smooth turns. 
    \item \emph{MANEUVERING}: Such episodes also concern changes in heading, but smaller ones that denote minor corrections in the course of the vessel or local maneuvers at ports. If the overall change in heading is less than $\phi$ (e.g., $5^{o}$), such episodes are marked as maneuvers; otherwise, they represent more significant turns.
    \item \emph{COMMUNICATION GAP}: As already discussed, this concerns gap intervals lasting less than $\delta$ time units. Since the vessel's whereabouts are unknown during such a gap, this episode is abstracted by the known start and end locations of the gap. 
    \item \emph{SAILING}: This characterizes every other segment where the vessel is considered moving, i.e., it is neither stopped nor turning and not in communication gap. As the vessel typically follows a straight course during this episode, it can be abstracted by its two endpoints.
\end{itemize}

For each episode, several \emph{statistics} can be calculated, all based on originally reported AIS locations. As illustrated within the green box in Figure~\ref{fig:example-trip-csv}, these include duration (in seconds), travelled distance (in nautical miles), as well as average speed (in knots) during this episode. They also include the total change in heading (in degrees) and direction (e.g., south west, north) when the vessel is turning or maneuvering; in all other episodes, such values indicate the overall heading and general direction.

\subsection{Context Enrichment}
\label{sec:context}

While segmenting long trajectories into trips and episodes can certainly facilitate their processing, such representations can be further enhanced with contextual information in order to provide deep insights into the motion patterns of vessels. Our workflow takes advantage of open data sources such as OpenStreetMap (OSM)\footnote{\href{https://www.openstreetmap.org/}{https://www.openstreetmap.org/}}, OpenSeaMap\footnote{\href{https://map.openseamap.org/}{https://map.openseamap.org/}} or the World Port Index\footnote{\href{https://msi.nga.mil/Publications/WPI}{https://msi.nga.mil/Publications/WPI}} to extract coastal or offshore \emph{geospatial}  features using topological checks (e.g., proximity to ports or crossing environmentally sensitive areas) along or near a vessel's course and ingest them into contextual information per episode.
Additional context is obtained from \emph{array-oriented scientific data} (typically in NetCDF format\footnote{\href{https://www.unidata.ucar.edu/software/netcdf}{https://www.unidata.ucar.edu/software/netcdf}}). This latter data (at varying grid resolutions) may concern \emph{weather conditions} available from large meteorological organizations such as Copernicus\footnote{\href{https://cds.climate.copernicus.eu/datasets}{https://cds.climate.copernicus.eu/datasets}}or NOAA\footnote{\href{https://www.noaa.gov/}{https://www.noaa.gov/}}, as well as \emph{bathymetry} data from GEBCO\footnote{\href{https://www.gebco.net/}{https://www.gebco.net/}}. Such information is extremely important for maritime situational awareness and can help explain motion patterns, e.g., a trip detour caused by a storm, or diverse courses taken by vessels of different draught due to shallow waters.

More specifically, our workflow is currently able to ingest extra context \emph{per episode} regarding:

\begin{itemize}
    \item \emph{Ports}: Identifies the name of the port where the vessel is anchored during a stop episode.
    
    \item \emph{Coastal features}: Finds whether the vessel is moving in close distance (e.g., less than 5 nautical miles) to capes, peninsulae, straits, etc. 
    
    \item \emph{Offshore areas}: Finds any polygonal regions (e.g., marine protection zones, national sea parks, fishing areas) the vessel is crossing along its course.

    \item \emph{Traffic separation schemes}: Indicates whether the vessel is navigating across the designated lane for its direction in high-density areas according to traffic regulations.

    \item \emph{Meteorological}: If such data is available for this time interval, it identifies the wind conditions (wind force in the Beaufort scale, wind direction) along the polyline that represents the course of a vessel during this episode. 
   
    \item \emph{Bathymetry}: Provides the seabed depth (in meters) at the grid cell of the stop event (if the vessel is anchored) or the minimum depth along the polyline of a moving episode (i.e., when the vessel is sailing or turning).
\end{itemize}

Any context obtained from geospatial features is listed as a collection of \emph{placemarks} in the resulting representation, as shown within the red box in Figure~\ref{fig:example-trip-csv}. Context features from NetCDF data on weather and bathymetry are maintained as separate items (e.g., \emph{wind\_beaufort}, \emph{sea\_depth}). 

Finally, more contextual information may be added concerning the \emph{entire trip}. For instance, in case that both the \emph{origin} and \emph{destination ports} have been identified (i.e., in the first and last episode of a trip), the entire trip geometry can be checked for possible match against known routes (e.g., \emph{ferry lines}  available from OSM). In case of match, \emph{Dynamic Time Warping} ($DTW$) can be used to measure similarity and thus highlight any significant deviations of the current trip from the typical ferry route. For example, we can identify that the vessel in Figure~\ref{fig:trip-annotated-example} makes a significant detour from the typical route (shown in blue dashed line) since the estimated $DTW =$ 2.91 is significantly high.

\subsection{Output Representations}
\label{sec:output}


\begin{figure}[!t]
\begin{subfigure}[t]{0.5\linewidth}

    \begin{center}
    \fbox{
    \begin{minipage}{0.9\linewidth}
    \scriptsize\ttfamily
    \{"direction":"SOUTH WEST",\\
    \quad "heading": -144,\\
    \quad "duration\_seconds": 1176,\\
    \quad "distance\_miles": 4.64,\\
    \quad "speed\_knots": 14.2,\\
    \quad "movement": "SAILING",\\
    \quad "wind\_beaufort": 4,\\
    \quad "wind\_direction": "WEST",\\
    \quad "sea\_depth": -11,\\
    \quad "placemarks": "off Lolland crossing Fehmarnbelt",\\
    \quad "start\_location": \{...\},\\
    \quad "end\_location": \{...\} 
    \}
    \end{minipage}
    }
    \end{center}
\vspace{-3mm}    
\caption{JSON}\label{fig:episode_json}
\end{subfigure}
\begin{subfigure}[t]{0.48\linewidth}
\begin{center}
    \fbox{
    \begin{minipage}{0.9\linewidth}
    \scriptsize\ttfamily
    From 2024-01-23 22:51:17 until 2024-01-23 23:10:52 SAILING SOUTH WEST at -144 degrees off Lolland crossing Fehmarnbelt, covering 4.64 nautical miles in 1176 seconds at speed 14.2 knots at minimum sea depth of -11 meters while WEST winds of intensity 4B were blowing. \vspace{3.4mm}
    \end{minipage}
    }
    \end{center}
    \vspace{-3mm}
\caption{TXT}\label{fig:episode_txt}
\end{subfigure}
    \caption{Alternative episode representations.}
\label{fig:episode}
\end{figure}

\begin{figure*}[!t]
\small  
\begin{promptbox}
\noindent\textbf{System:} 

\noindent\textbf{\em ROLE AND OBJECTIVE}: You are a data scientist who studies trips of vessels, i.e., the route of ships sailing between ports or locations at open sea. You are given a list of waypoints in JSON format representing the trip of a vessel, including information about anchorage in ports, duration of each stop, sailing for varying time intervals, and occasionally turning maneuvers when a vessel changes its direction of movement. Based on this information about the trip, you must provide its description in plain text. 

\vspace{0.25cm}

\noindent\textbf{\em APPROACH AND LIMITATIONS:}

(1) All vessel locations are specified in (longitude, latitude) coordinates in WGS84 (EPSG:4326) Coordinate Reference System. 

(2) Distances are always expressed in nautical miles and are calculated using the Haversine formula on an ellipsoid. 

(3) Speed values are in knots, i.e., nautical miles per hour. 

(4) Durations are in seconds. All timestamp values are in UNIX epochs (seconds elapsed since 1970-01-01) in GMT time zone. Wind intensity is in Beaufort scale. 

(5) Placemarks indicate proximity to islands, ports, straits, capes, etc. 

(6) Note that the motion patterns and speed of vessels absolutely depend on their type (e.g., Passenger, Cargo, Tanker, Pleasure craft, etc.), and some vessels may be anchored for long intervals. 

(7) Due to COMMUNICATION GAPS, a vessel may suddenly appear far away from its previously known position.

\vspace{0.25cm}

\noindent\textbf{\em METHODOLOGY}: You must provide:

(i) A textual description in PLAIN TEXT that outlines the trip. In a separate paragraph describe each major stage of the trip, such as port operations (e.g., anchorage), important navigational features (e.g., cruising at open sea, maneuvering, turning) as well as any proximity to placemarks (e.g., islands, straits, capes). Do NOT include any explanations or reasoning regarding any stage of the trip.

(ii) A short summary in JSON format with statistics that include the traveled distance (in nautical miles), the total duration (in seconds), the names of the origin and destination ports (if available) and any adverse weather conditions identified across the entire trip as follows:

\texttt{\textcolor{blue}{\{"traveled\_distance": distance\_value, "total\_duration": duration\_value,
"origin\_port": port\_name,
"destination\_port" : port\_name,
"adverse\_weather\_conditions": [ "wind\_intensity":"beaufort\_value", "wind\_direction": direction\_value ]
\}}}

For your responses, you must NOT include information from any external source; use information strictly from the input JSON.

\vspace{0.5cm}

\noindent\textbf{User:} 

Based on this JSON array representing the simplified trip of a vessel, please provide a textual description. \texttt{\textcolor{purple} {```json\{ ... \}```}}

\end{promptbox}
\vspace{-0.2cm}
\caption{\normalfont Prompt template for generating trajectory descriptions. The LLM receives as input a \texttt{\textcolor{purple}{json}} \textcolor{purple}{array} with distilled context-enhanced representation about a vessel trip and generates: (i) semantically-rich natural language descriptions in plain text, and (ii) overall trip statistics (in \texttt{\textcolor{blue}{json}} \textcolor{blue}{format}).
}
\label{fig:trajectory-description-generator-prompt}
\end{figure*}

The resulting contextually enriched semantic trajectories can be exported in several output formats:

\begin{itemize}
    \item \emph{CSV}: A separate row is emitted per episode using a predefined attribute schema. 
    The start and end \emph{point} locations of each episode are listed in separate columns in Well-Known Text (WKT).
    
    \item \emph{MAP}: This representation is also in CSV format, but it includes a single \emph{geometry} column to facilitate map visualization (e.g., in GIS software). Each episode is represented with a linestring geometry in WKT (like the ones shown in Figure~\ref{fig:example-trip-csv}), which in the case of turns or maneuvers includes a sequence of vertices for more accurate illustration of such events. 

    \item \emph{JSON}: Each trip is serialized as a JSON array of elements, each one representing an episode. As shown in Figure~\ref{fig:episode_json}, an episode is described through a set of key–value pairs capturing the mobility characteristics and contextual information associated with that segment.
    \item \emph{TXT}: An entire trip is summarized into a paragraph in plain text. 
    We stress that this is {\bf not} an AI-generated narrative (we discuss in the next 
    section how this can be achieved), but rather a tedious yet factual recitation of 
    all available mobility and contextual information using standard syntax and 
    predefined connecting words. As illustrated in the example in Figure~\ref{fig:episode_txt}, an episode is described by a separate sentence that simply replicates all extracted features of the episode verbatim.
    
\end{itemize}
\section{LLM-Generated Descriptions of Vessel Trips}
\label{sec:generator}

\begin{figure*}[!t] 
\begin{subfigure}[t]{0.48\linewidth}
\centering
    \includegraphics[width=0.9\linewidth]{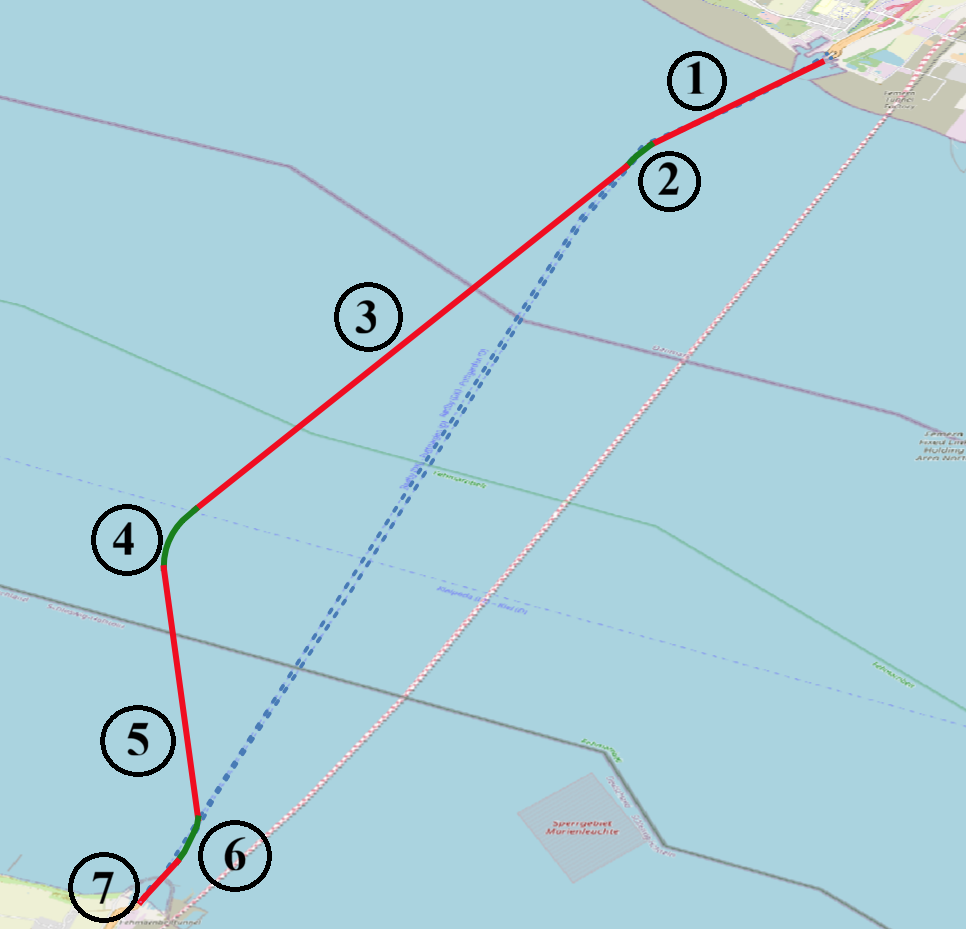}
    \caption{Map illustration of input trajectory.}
\end{subfigure}
\begin{subfigure}[t]{0.5\linewidth}
\begin{enumhighlight}

\begin{enumerate}[
    label=\circled{\arabic*},
    leftmargin=*,
    labelsep=0.6em
]  
    \item The vessel began its voyage off the coast of Lolland, sailing south-west at about 9 knots for roughly 1.4 NM.  
    \item It then performed a short turning manoeuvre still off Lolland, covering a small distance while increasing speed to around 11 knots. 
    \item Continuing south-west, the ship crossed the Fehmarn Belt under a moderate west-westerly breeze (Beaufort~4), sailing at 14 knots for more than 4 NM.  
    \item A brief turn to the south was made within the belt.
    \item A steady southward leg followed at 15 knots, covering about 2.7 NM while the same moderate wind persisted.  
    \item The vessel executed another turning manoeuvre to the south-west. Then proceeded at reduced speed (6.5~knots) toward the approach of Putgarden.
    \item A final short, very slow leg (1 knot) brought the ship to the anchorage area in Putgarden port where the vessel stopped for over an hour.
\end{enumerate}
\end{enumhighlight} 
\caption{LLM-generated textual description using openai/gpt-oss-120b.}
\end{subfigure}
    \caption{Trajectory episodes (depicted on map with alternating colors) given as input to LLM for generating a textual description.}
\label{fig:description}
\end{figure*}

Leveraging the powerful capabilities of LLMs to
extract tokens directly from continuous spatio-temporal data (like AIS) is non-trivial and is sensitive to prompt design; as a result, such trajectory descriptions generated by LLMs are not always grounded on actual information \cite{xu2025trajectorypredictionmeetslarge}.
Instead, we propose to generate descriptions in natural language strictly based on contextually-enriched trajectory representations as issued from our workflow. Such descriptions can enable advanced mobility analysis and semantic reasoning in downstream trajectory tasks (e.g., forecasting, imputation, anomaly detection). 



To generate descriptions using LLMs, we construct a prompt (Figure~\ref{fig:trajectory-description-generator-prompt}) consisting of a system message that establishes the task context and a user message that provides the extracted trajectory information. 
This approach includes detailed instructions concerning interpretation of the various features (coordinates, timestamp, speed, distance, duration) per episode in the movement, as well as precise guidelines for measurement units (e.g., nautical miles for distance, Beaufort scale for wind force) and any available placemarks (islands, ports, capes, etc.) added from extra context to make sure that the model can interpret and use them properly. Most importantly, specific motion patterns for vessels (e.g., anchorage for long periods of time, possible communication gaps en route) are highlighted.

This prompt also includes a methodology that guides the LLM in synthesizing available information in order to provide its response in two parts:

\begin{itemize}
    \item A {\em textual description} in natural language that outlines the trip and its most significant stages. Strictly based on the input trajectory representation in episodes (in JSON format), the LLM should identify port operations (e.g., anchorage), important navigational features (e.g., cruising at open sea, maneuvers, turns) as well as any proximity to placemarks (e.g., islands, straits, capes).
    \item A summary of {\em trip statistics} in JSON format that include the traveled distance, the total duration, and the origin and destination ports (if available). Thus, the LLM must aggregate such features from the input representation, making the necessary calculations (e.g., adding distances, averaging speed values) across the episodes of the trip. In case of any adverse weather conditions identified along the entire trip, these should be also included as well.
\end{itemize}

The textual description per episode listed in Figure~\ref{fig:description} (concerning the same trajectory depicted in Figure~\ref{fig:trip-annotated-example}) is indicative of the LLM responses. Comparing the obtained description of episode \circled{3} with its respective input representations shown in Figure~\ref{fig:episode}, observe how intelligently the LLM grasps information and conveys the vessel's movement. Strictly based on the given semantic trajectory representations concerning mobility, geospatial, and context features, LLMs seem capable to synthesize a succinct summary that captures the successive episodes of each trip and highlight essential information of each motion path.

\section{Empirical Evaluation}
\label{sec:evaluation}

We have conducted an empirical study by generating descriptions in natural language for contextually-enriched vessel trip representations reconstructed from real-world AIS data. We not only report performance results regarding contextualization and trip statistics, but we also assess the quality of textual descriptions generated by different LLMs. These latter tests evaluate the ability of AI models to understand mobility and spatiotemporal features pertinent to vessel trajectories and issue a succinct, yet factual account of each trip.

\subsection{Experimental Setup}
\label{sec:setup}

\stitle{Datasets.}
We extracted raw AIS positions regarding \emph{passenger} ships for a 3-month period (January--March 2024) available by the Danish Maritime Authority\footnote{\href{https://www.dma.dk/safety-at-sea/navigational-information/ais-data}{https://www.dma.dk/safety-at-sea/navigational-information/ais-data}}. For \emph{context} enrichment, we extracted geospatial datasets for this region from OpenStreetMap (protected areas, capes, peninsulae, straits, islands, ferry lines), OpenSeaMap (traffic separation schemes), and World Port Index (ports). We also used NetCDF sources, namely meteorological data from Copernicus (wind features), and bathymetry features from GEBCO (sea depth).

We applied our workflow to extract context-enriched trajectory representations in JSON format. Finally, we randomly selected 460 trips between 10 specific ports in the region to be given as input to each LLM and generate their descriptions. The selected trips represent 212,203 raw AIS locations.

\stitle{Models.}
We tested several open LLMs of varying capabilities (\emph{llama-3.1-8B-instant, llama-3.3-70B-versatile, qwen-3-32B, gpt-oss-20B, gpt-oss-120B}) in generating textual descriptions.
All LLMs are queried with a temperature $0.1$ to reduce output variance. All requests are issued through Groq API\footnote{\href{https://groq.com/}{https://groq.com/}} using fixed model versions and the same prompt (Figure~\ref{fig:trajectory-description-generator-prompt}) to ensure a fair treatment.

\subsection{Performance Results}
\label{sec:performance}

\begin{figure}[t]
    \centering
    \includegraphics[width=0.8\linewidth]{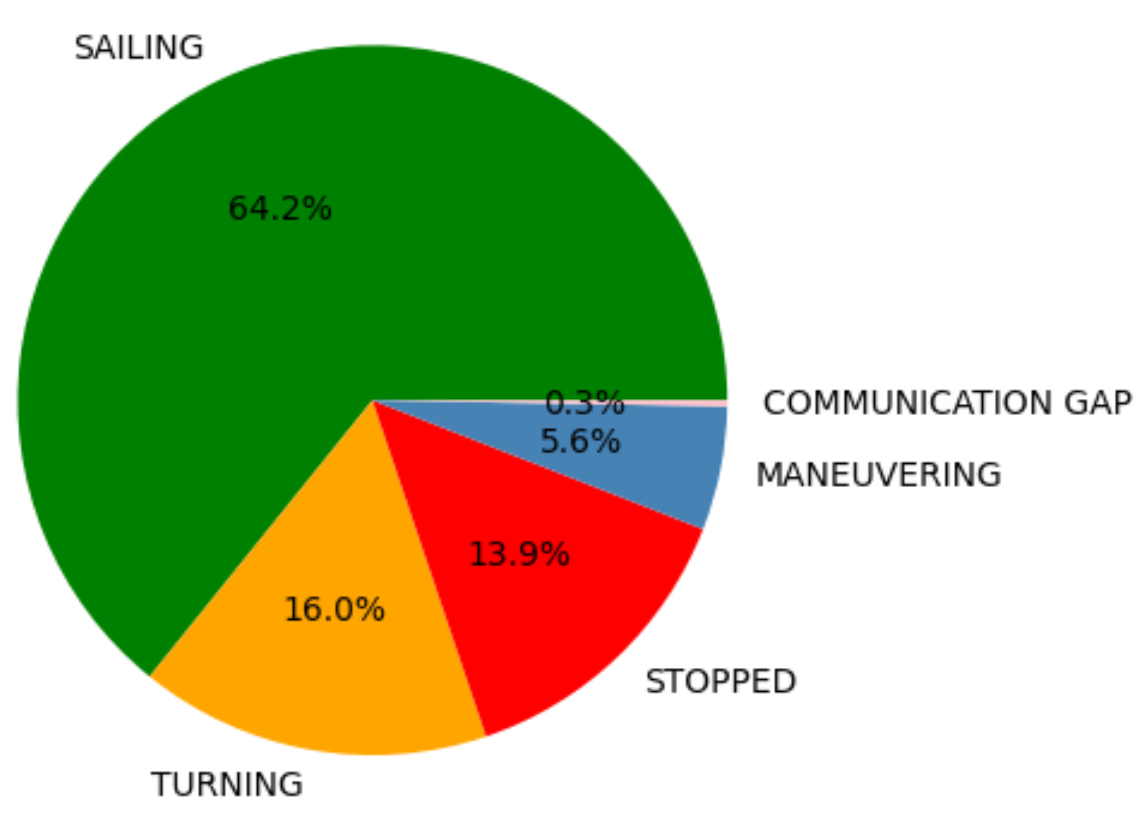}
     \caption{Distribution of annotated episodes among all trips.}
    \label{fig:distr_episodes}
\end{figure}

\stitle{Context-enriched semantic trajectories.}
We first examine features in semantic trajectories after applying our processing framework (Section~\ref{sec:methodology}). Note that this does not include interaction with LLMs. Figure~\ref{fig:distr_episodes} illustrates the breakdown (\%) of the type of episodes among all 460 trips detected in the AIS dataset. In total, 3,332 episodes were identified, and most of them represent sailing legs in trips. A significant number (over 21\%) of episodes also concerns turns or maneuvers, while stop episodes have a fair share (13.9\%). Note that very few episodes concern communication gaps in a trip; recall that each such gap spans at most $\delta=1$ hour, because larger gap durations are used to split the trajectory into trips. 

\begin{figure}[t]
    \centering
    \includegraphics[width=0.85\linewidth]{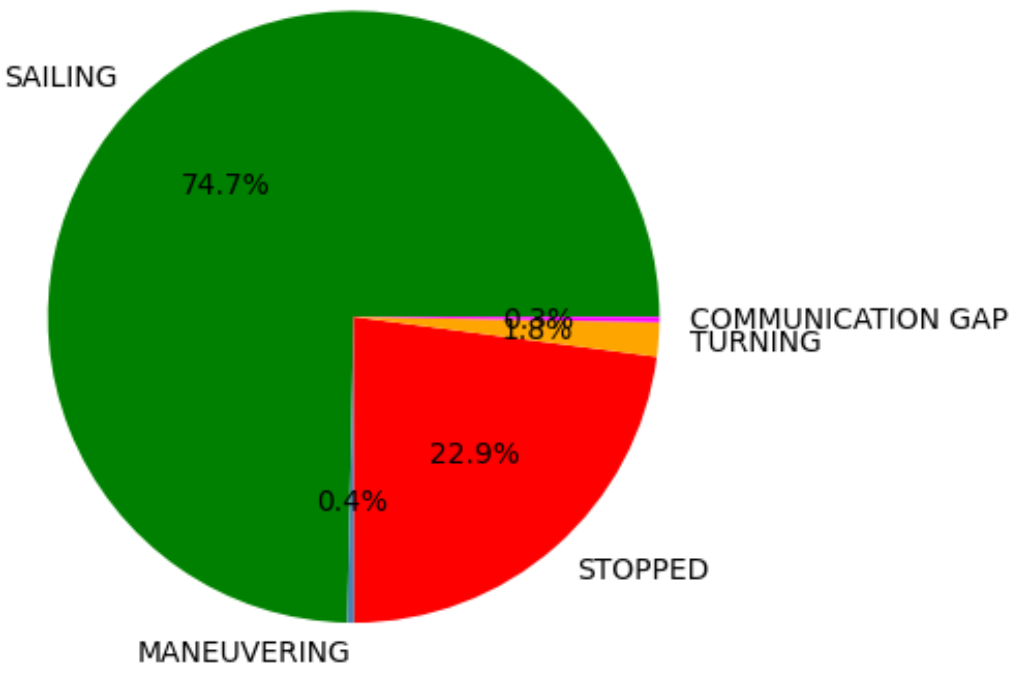}
     \caption{Cumulative duration of annotated episodes in all trips.}
    \label{fig:duration_episodes}
\end{figure}

However, when considering the cumulative duration of each type of episodes, we get another interesting insight. As illustrated in Figure~\ref{fig:duration_episodes}, the sailing legs and the stop periods cover most of the duration in the examined trips. Only about 2\% of total trip time is spent in turns or maneuvers, although these represent more than one fifth of the detected episodes (Figure~\ref{fig:distr_episodes}). The total duration of gaps is almost negligible, given that only rarely a vessel is lost during a trip in this area.

Regarding enrichment with extra context, Table~\ref{tab:context-episodes} lists the percentage in each type of episodes that obtained such contextual information. Statistics are shown for each category of sources, i.e., \emph{geospatial} (ports, straits, capes, etc.) mainly extracted from OSM, \emph{weather} (meteorological NetCDF files), and \emph{bathymetry} (also NetCDF data). Note that a large portion of moving episodes (i.e.,  sailing, turning, or maneuvering) is associated with spatial features (e.g., proximity to capes). Most importantly, the respective port is identified for all stops. Weather data was available for a significant part of moving episodes, whereas bathymetry data was associated for the vast majority of sailing and turning legs of trips.
Of course, no context can be associated with communication gaps, as vessel whereabouts are unknown.  

\begin{table}[t]
\centering
\renewcommand{\arraystretch}{1.2}
\caption{\normalfont Percentage (\%) of episodes enriched with context.}
\label{tab:context-episodes} 
\small
\begin{tabular}{l||c|c|c}
\hline
\textbf{Episode type} & geospatial & weather & bythometry \\
\hline
\hline
SAILING & 63 & 39 & 80 \\
TURNING & 54 & 37 & 98 \\
MANEUVERING & 41 & 12 & 64 \\
STOPPED & 100 & 0 & 37 \\
COMMUNICATION GAP & 0 & 0 & 0 \\
\hline
\end{tabular}
\end{table}

\stitle{Generating trip descriptions with LLMs.}
Based on semantically enriched trajectory representations, we employed several LLMs to generate descriptions in natural language per trip. Next, we report performance statistics, deferring assessment of their quality for Section~\ref{sec:assessment}.

Table~\ref{tab:llm-response-time} depicts the average time required by each LLM in order to return a trip description. Apparently, smaller models can offer their response in about a second; larger models take longer, but still manage to answer within a few seconds on average, which highlights their potential for practical use by maritime stakeholders

\begin{table}[t]
\centering
\renewcommand{\arraystretch}{1.2}
\caption{\normalfont Average response time in generating a trip description using different LLMs.}
\label{tab:llm-response-time} 
\small
\begin{tabular}{l||c}
\hline
\textbf{LLM Model} & \textbf{ response time (sec)} \\
\hline
\hline
llama-3.1-8b-instant & 1.01 \\

llama-3.3-70b-versatile & 1.12 \\

qwen/qwen3-32b & 2.58 \\

openai/gpt-oss-20b & 1.45 \\

openai/gpt-oss-120b & 2.36 \\
\hline
\end{tabular}
\end{table}

\begin{figure*}[!t]
\begin{promptbox}
\small  

\noindent\textbf{System:} 

\noindent\textbf{\em ROLE AND OBJECTIVE}: You are a data scientist who studies trips of vessels, i.e., the route of ships sailing between ports or locations at open sea. You are given AI-generated textual descriptions of such trips, including information about anchorage in ports, duration of each stop, sailing for varying time intervals, and occasionally turning maneuvers when a vessel changes its direction of movement. You must evaluate the quality of these descriptions on factuality, accuracy, and comprehensiveness with respect to the original input information. 

\vspace{0.25cm}

\noindent\textbf{\em APPROACH AND LIMITATIONS:}

Each description in PLAIN TEXT outlines the trip of a vessel. In a separate paragraph it describes each major stage of the trip, such as port operations (e.g., anchorage), important navigational features (e.g., cruising at open sea, maneuvering, turning) as well as any proximity to placemarks (e.g., islands, straits, capes). In generating each description, the model was asked NOT to include information from any external source, but only use information strictly from the input JSON. The description does NOT include any explanations or reasoning regarding any stage of the trip. 

(1) All vessel locations are specified in (longitude, latitude) coordinates in WGS84 (EPSG:4326) Coordinate Reference System. 

(2) Distances are always expressed in nautical miles and are calculated using the Haversine formula on an ellipsoid. 

(3) Speed values are in knots, i.e., nautical miles per hour. 

(4) Durations are in seconds. Wind intensity is in Beaufort scale. 

(5) Placemarks indicate proximity to islands, ports, straits, capes, etc.

(6) Note that the motion patterns and speed of vessels absolutely depend on their type (e.g., Passenger, Cargo, Tanker, Pleasure craft, etc.), and some vessels may be anchored for long intervals. 

(7) Due to COMMUNICATION GAPS, a vessel may suddenly appear far away from its previously known position.

\vspace{0.25cm}

\noindent\textbf{\em EVALUATION METHODOLOGY}: 

Evaluate the description in each one of the following criteria:

(i) \emph{Relevance}: Did it actually address the user's request and provided a reliable account of the trip according to the specified guidelines? 

(ii) \emph{Faithfulness}: Did it stick to the facts stipulated in the input instead of wandering into hallucinations?

(iii) \emph{Correctness}: Was it accurate from start to finish or discrepancies (overestimations or underestimations) are observed in several measurements (e.g., distance, speed, duration) along the trip?

Please rate this description with scores on a 1-5 scale for each quality criterion:

{\bf 1} - \emph{Poor}: description below expectations with several errors and inconsistencies.

{\bf 2} - \emph{Unsatisfactory}: description partially meets requirements, occasionally missing important aspects of the trip.

{\bf 3} - \emph{Fair}: description meets requirements offering a consistent account of the entire trip.

{\bf 4} - \emph{Very good}: high quality, coherent description.

{\bf 5} - \emph{Excellent}: description significantly exceeds all expectations.

\noindent\textbf{\em EVALUATION RESULT}: 

You must provide your evaluation scores as well as a brief explanation of your scoring in JSON format, as in the following example:

\texttt{\textcolor{blue}{\{"relevance\_score": 4, "faithfulness\_score": 5, "correctness\_score": 3, "explanation": "The description was clear and comprehensive, accurately covering all stages of the trip but certain measurements on distance and speed deviated from actual values."\}}}

\vspace{0.5cm}

\noindent\textbf{User:}

\noindent\textbf{\em INPUT}: This JSON array represents the trip of a vessel:
\texttt{\textcolor{purple} {```json\{ ... \}```}}
 
\noindent\textbf{\em OUTPUT}: This is the AI-generated textual description of this trip:
\texttt{\textcolor{green} {"{NL DESCRIPTION}"}} 

\noindent\textbf{\em TASK}: Evaluate the quality of the output description in terms of relevance, faithfulness, and correctness with respect to the input JSON.

\end{promptbox}
\vspace{-0.2cm}
\caption{\normalfont Prompt template for evaluating the generated trajectory descriptions. The LLM receives as input a \textcolor{purple}{\texttt{json} array} with distilled context-enhanced representation about a vessel trip and evaluates the quality of the \texttt{\textcolor{green}{generated NL description}} in terms of relevance, faithfulness, and correctness.
}
\label{fig:trajectory-description-evaluation-prompt}
 \end{figure*}

As each response also includes \emph{trip statistics} (the second output requested from LLMs), we also evaluated the accuracy of the returned estimates. 
We report statistics (mean, standard deviation, maximum) over \emph{deviations} (\%) in \emph{total trip duration} (Table~\ref{tab:llm-duration_deviation}) and \emph{travelled distance} (Table~\ref{tab:llm-distance_deviation}) from the respective ground truth values (computed over original AIS locations per trip). Results in \textbf{bold} indicate LLMs with the best performance. Despite specific prompt instructions on how to handle spatiotemporal measurements and units,  estimates from models may diverge from actual values. Unsurprisingly, smaller models (like \emph{llama-3.1-8B-instant}) seem less capable in accurately estimating trip statistics and incur very significant over- or underestimations. In contrast, larger models (like \emph{openai/gpt-oss-120b}) offer much better estimates, although sometimes they can also be off the actual value by orders of magnitude (as indicated by the max \% deviation). This finding underscores current limitations of LLMs in conducting accurate calculations, especially concerning mobility features crucial for trajectory analysis.

\begin{table}[t]
\centering
\renewcommand{\arraystretch}{1.2}
\caption{\normalfont Statistics on {\em deviations} (\%) of {\em trip durations} estimated by LLM models w.r.t. ground truth.}
\label{tab:llm-duration_deviation} 
\small
\begin{tabular}{l||r|r|r}
\hline
\textbf{LLM Model} & \textbf{mean} & \textbf{stdev} & \textbf{max} \\
\hline
\hline
llama-3.1-8b-instant   & 113.72 & 1047.73 & 19017.24 \\
llama-3.3-70b-versatile   & 13.68 & 16.79 & {\bf 85.00} \\
qwen/qwen3-32b   & 7.86 & 17.88 & 206.38 \\
openai/gpt-oss-20b   & 1.87 & 6.74 & 88.10 \\
openai/gpt-oss-120b   & {\bf 1.53} & {\bf 6.47} & 88.10 \\
\hline
\end{tabular}
\end{table}

\begin{table}[t]
\centering
\renewcommand{\arraystretch}{1.2}
\caption{\normalfont Statistics on {\em deviations} (\%) of {\em travelled distance} per trip as estimated by LLM models w.r.t. ground truth.}
\label{tab:llm-distance_deviation} 
\small
\setlength{\tabcolsep}{8pt}
\begin{tabular}{l||r|r|r}
\hline
\textbf{LLM Model} & \textbf{mean} & \textbf{stdev} & \textbf{max} \\
\hline
\hline
llama-3.1-8b-instant   & 13.79 & 26.36 & 248.28 \\
llama-3.3-70b-versatile   & 2.44 & 4.52 & {\bf 34.55}  \\
qwen/qwen3-32b    & 4.92 & 43.48 & 848.22 \\
openai/gpt-oss-20b   & 1.06 & 4.29 &  35.32 \\
openai/gpt-oss-120b    & {\bf 0.89} & {\bf 4.12} & 35.32 \\
\hline
\end{tabular}
\end{table}

\subsection{Qualitative Assessment of Generated Descriptions}
\label{sec:assessment}

\begin{table}[t]
\centering
\renewcommand{\arraystretch}{1.2}
\caption{\normalfont Qualitative evaluation of textual descriptions of vessel trips generated by diverse LLM models.}
\label{tab:llm-evaluation} 
\small
\begin{tabular}{l||c|c|c}
\hline
\textbf{LLM Model} & \textbf{relevance} & \textbf{faithfulness} & \textbf{correctness} \\
\hline
\hline
llama-3.1-8b-instant   & 3.90 &	3.84 & 3.27 \\
llama-3.3-70b-versatile   & 4.47 & 4.77 & 4.36  \\
qwen/qwen3-32b    & 4.42 & 4.48 & 3.82 \\
openai/gpt-oss-20b   & 4.69 &	4.81 &	4.35 \\
openai/gpt-oss-120b    & {\bf 4.91} &	{\bf 4.96} & {\bf 4.69} \\
\hline
\end{tabular}
\end{table}

Evaluating the quality of such LLM-generated narratives is a challenging task due to absence of ground truth. To tackle this, we opted to employ an LLM as evaluator, according to the emerging paradigm of \emph{"LLM-as-a-Judge"}~\cite{GU2026101253}. To ensure that such evaluation would be as objective, consistent, and reliable as possible, we employed a different LLM from another AI provider, namely \emph{gemini-2.5-flash}\footnote{\href{https://ai.google.dev/gemini-api/docs/models/gemini-2.5-flash}{https://ai.google.dev/gemini-api/docs/models/gemini-2.5-flash}} using in-context learning. More specifically, we designed another prompt specifically for this evaluation (shown in Figure~\ref{fig:trajectory-description-evaluation-prompt}) in order to assess the quality of each generated description individually. The evaluator is given the same input representation of a trip (i.e., a context-enriched semantic trajectory in JSON format) with the same guidelines for generating descriptions. An LLM-generated description (using one of the five models mentioned in Section~\ref{sec:setup}, \emph{unknown} to the evaluator) is also given, and the evaluator is asked to rate its quality using three \emph{evaluation criteria} (aka \emph{dimensions}~\cite{GU2026101253}):

\begin{itemize}
    \item \emph{Relevance} rates whether the description offers a relevant response by giving a reliable overall account of the trip according to the stipulated guidelines.
    \item \emph{Faithfulness} assesses whether the description is actually grounded on the facts given in the input JSON representation without wandering into hallucinations.
    \item \emph{Correctness} evaluates the accuracy of the entire description and examines if discrepancies (over- or underestimations) are observed in measurements (e.g., distance, speed, duration) along the trip.
\end{itemize}

The evaluator assigns \emph{scores} to each description according to the aforementioned dimensions. Each dimension is scored on a scale of 1 to 5, ranging from worst to best:
\begin{enumerate}
\item \emph{Poor}: The description contains many errors (e.g., misinterpretation of values or units in duration of an episode) or factual inconsistencies (e.g., ignoring a leg of the trip). This score rates such severely distorted descriptions below expectations. 


\item \emph{Unsatisfactory}: Major discrepancies in stated values are observed or some important aspects of the trip may occasionally be missed. Thus, the description only partially meets requirements and cannot be considered reliable.

\item \emph{Fair}: No significant errors are found, and all legs of the trip are mentioned. Such a description meets requirements, offering a consistent account of the entire trip.

\item \emph{Very good}: This is a high-quality, factual, coherent, and comprehensive description of the trip without bias.

\item \emph{Excellent}: The description significantly exceeds all expectations. It creatively compiles all available information into a fluent, accurate, and reliable trip narrative.
\end{enumerate}

The evaluator is also asked to justify its scores with an \emph{explanation}; this can offer insight into the assessment process.

We performed this qualitative evaluation of all trip descriptions generated by the five models (in total, 2300 prompt requests). Table~\ref{tab:llm-evaluation} reports averages in each score per LLM. Unsurprisingly, the most powerful model (\emph{openai/gpt-oss-120b}) achieves the finest descriptions by all three criteria. With very few exceptions, it sticks to the input facts, and rarely its reported measurements deviate from the actual ones. In contrast, smaller models fall behind significantly, especially with respect to correctness. For instance, \emph{lama-3.1-8b-instant} often exaggerated by orders of magnitude in its reported values on speed or duration of many episodes, due to its limited calculation capabilities as mentioned before. Other models with greater capacity to learn (like \emph{llama-3.3-70b-versatile}, \emph{qwen/qwen3-32b}, and \emph{openai/gpt-oss-20b}) offer very good responses with factual narratives, but occasional inaccuracies in measurements. Again, this finding indicates the current limitations of LLMs in interpreting and correctly processing even simple spatiotemporal features like distance or speed.


\section{Concluding Remarks}
\label{sec:conclusion}

In this paper, we proposed a methodology for constructing contextually-enriched representations of vessel trajectories collected from AIS messages. By semantically annotating important locations along each vessel's course that signify mobility events like stops, turns, or communication gaps, we are able to segment long trajectories into distinct trips, each one consisting of a sequence of episodes. Geographical and maritime features from open sources (e.g., ports, capes, islands, or traffic separation zones), as well as meteorological and bathymetry data can add rich contextual information to each episode, providing important insights into vessel activity. Finally, such trajectory representations can be given as input to powerful LLMs in order to generate  descriptions in natural language that effectively summarize entire trips (each possibly spanning thousands of raw AIS locations) in just a few sentences and highlight their most important mobility patterns.

We plan to improve such descriptions by taking into account vessel interactions across their trips (i.e., not each trajectory in isolation), offering richer context crucial for maritime safety. Utilizing such descriptions can be advantageous in downstream tasks, like trajectory forecasting for optimized navigation, imputation of trajectory segments missed due to gaps, or anomaly detection (e.g., identify detours), etc.

\vspace{0.3cm}
\noindent{\bf Acknowledgments.}
This work has been partially supported by project MIS 5154714 of the National Recovery and Resilience Plan Greece 2.0 funded by the European Union under the NextGenerationEU Program.

\bibliographystyle{ieeetr}
\bibliography{references}

@String { PVLDB      = {{PVLDB}}}

@String { SIGSPATIAL = {{SIGSPATIAL}}}

@String { TSAS       = {{TSAS}}}

@String { KDD        = {{KDD}}}

@String { EDBT        = {{EDBT}}}

@String { PVLDB      = {VLDB Endowment ({PVLDB})}}

@String { SIGSPATIAL = {{ACM Intl. Conf. on Advances in Geographic Information Systems (SIGSPATIAL)}}}

@String { TSAS       = {{ACM Trans. Spatial Algorithms Syst. (TSAS)}}}

@String { KDD        = {{ACM} Intl. Conf. on Knowledge Discovery and Data Mining ({KDD})}}

@String { EDBT       = {Intl. Conf. on Extending Database Technology ({EDBT})}}

@article{10.1007/s10707-022-00475-0,
title = {{Optimizing Vessel Trajectory Compression for Maritime Situational Awareness}},
  author = {Fikioris, Giannis and Patroumpas, Kostas and Artikis, Alexander and Pitsikalis, Manolis and Paliouras, Georgios},
  year = {2022},
  journal = {Geoinformatica},
  volume = {27},
  number = {3}
}

@article{yang2019big,
title = {{How Big Data Enriches Maritime Research--a Critical Review of Automatic Identification System (AIS) Data Applications}},
  author = {Yang, Dong and Wu, Lingxiao and Wang, Shuaian and Jia, Haiying and Li, Kevin X},
  year = {2019},
  journal = {Transport reviews},
  volume = {39},
  number = {6}
}

@article{atluri2018spatio,
title = {{Spatio-Temporal Data Mining: a Survey of Problems and Methods}},
  author = {Atluri, Gowtham and Karpatne, Anuj and Kumar, Vipin},
  year = {2018},
  journal = {ACM Computing Surveys},
  volume = {51},
  number = {4}
}

@article{musleh2023kamel,
title = {{Kamel: a Scalable Bert-Based System for Trajectory Imputation}},
  author = {Musleh, Mashaal and Mokbel, Mohamed F},
  year = {2023},
  journal = PVLDB,
  volume = {17},
  number = {3}
}

@article{musleh2024let,
title = {{Let's Speak Trajectories: a Vision to Use NLP Models for Trajectory Analysis Tasks}},
  author = {Musleh, Mashaal and Mokbel, Mohamed F},
  year = {2024},
journal = TSAS,
  volume = {10},
  number = {2}
}

@inproceedings{devlin2019bert,
title = {{BERT: Pre-Training of Deep Bidirectional Transformers for Language Understanding}},
  author = {Devlin, Jacob and Chang, Ming-Wei and Lee, Kenton and Toutanova, Kristina},
  year = {2019},
  booktitle = {Conf.  of the North American chapter of the association for computational linguistics}
}

@article{yang2025bert4traj,
title = {{BERT4Traj: Transformer Based Trajectory Reconstruction for Sparse Mobility Data}},
  author = {Yang, Hao and Yao, Angela and Whalen, Christopher and Mai, Gengchen},
  year = {2025},
  journal = { arXiv:2507.03062}
}

@article{si2023trajbert,
title = {{TrajBERT: BERT-Based Trajectory Recovery with Spatial-Temporal Refinement for Implicit Sparse Trajectories}},
  author = {Si, Junjun and Yang, Jin and Xiang, Yang and Wang, Hanqiu and Li, Li and Zhang, Rongqing and Tu, Bo and Chen, Xiangqun},
  year = {2023},
  journal = {IEEE Transactions on Mobile Computing},
  volume = {23},
  number = {5}
}

@inproceedings{liang2025foundation,
title = {{Foundation Models for Spatio-Temporal Data Science: a Tutorial and Survey}},
  author = {Liang, Yuxuan and Wen, Haomin and Xia, Yutong and Jin, Ming and Yang, Bin and Salim, Flora and Wen, Qingsong and Pan, Shirui and Cong, Gao},
  year = {2025},
  booktitle =KDD
}

@inproceedings{zhang2024large,
title = {{Large Language Models for Spatial Trajectory Patterns Mining}},
  author = {Zhang, Zheng and Amiri, Hossein and Liu, Zhenke and Zhao, Liang and Z{\"u}fle, Andreas},
  year = {2024},
  booktitle = {Workshop on Geospatial Anomaly Detection}
}

@article{wang2023would,
title = {{Where Would I Go Next? Large Language Models As Human Mobility Predictors}},
  author = {Wang, Xinglei and Fang, Meng and Zeng, Zichao and Cheng, Tao},
  year = {2023},
  journal = { arXiv:2308.15197}
}

@article{guo2025natural,
title = {{A Natural Language-Based Automatic Identification System Trajectory Query Approach Using Large Language Models}},
  author = {Guo, Xuan and Yu, Shutong and Zhang, Jinxue and Bi, Huanyu and Chen, Xiaohui and Liu, Junnan},
  year = {2025},
  journal = {ISPRS International Journal of Geo-Information},
  volume = {14},
  number = {5}
}

@article{merten2025using,
title = {{Using LLMs for Analyzing AIS Data}},
  author = {Merten, Gaspard and Dejaegere, Gilles and Sakr, Mahmoud},
  year = {2025},
  journal = { arXiv:2504.07557}
}

@article{ma2025navigation,
title = {{Navigation-GPT: A Robust and Adaptive Framework Utilizing Large Language Models for Navigation Applications}},
  author = {Ma, Feng and Wang, Xiu-min and Chen, Chen and Xu, Xiao-bin and Yan, Xin-ping},
  year = {2025},
  journal = { arXiv:2502.16402}
}

@article{liu2025vtllm,
title = {{VTLLM: A Vessel Trajectory Prediction Approach Based on Large Language Models}},
  author = {Liu, Ye and Xiong, Wei and Chen, Nanyu and Yang, Fei},
  year = {2025},
  journal = {Journal of Marine Science and Engineering},
  volume = {13},
  number = {9}
}

@article{chen2025msce,
title = {{MSCE: Empowering Vessel Identity Anomaly Detection with Multimodal LLMs}},
  author = {Chen, Nanyu and Yang, Anran and Chen, Luo and Wu, Hui and Jing, Ning},
  year = {2025},
  journal = {IEEE Transactions on Aerospace and Electronic Systems}
}

@inproceedings{mbuya2024trajectory,
title = {{Trajectory Anomaly Detection with Language Models}},
  author = {Mbuya, Jonathan Kabala and Pfoser, Dieter and Anastasopoulos, Antonios},
  year = {2024},
  booktitle = SIGSPATIAL
}

@inproceedings{you2025interpretable,
title = {{Interpretable Vessel Behavior Inference from AIS Trajectories via CoT-Enhanced LLMs}},
  author = {You, Zehao and Lin, Yinchen},
  year = {2025},
  booktitle = {Conf. on Big Data \& Artificial Intelligence \& Software Engineering}
}

@article{chen2025semint,
title = {{SEMINT: An LLM-Empowered Long-Term Vessel Trajectory Prediction Framework}},
  author = {Chen, Nanyu and Yang, Anran and Wu, Hui and Chen, Luo and Xiong, Wei and Jing, Ning},
  year = {2025},
  journal = {Journal of Geographical Information Science}
}

@article{park2025ais,
title = {{AIS-LLM: A Unified Framework for Maritime Trajectory Prediction, Anomaly Detection, and Collision Risk Assessment with Explainable Forecasting}},
  author = {Park, Hyobin and Jung, Jinwook and Seo, Minseok and Choi, Hyunsoo and Cho, Deukjae and Park, Sekil and Choi, Dong-Geol},
  year = {2025},
  journal = { arXiv:2508.07668}
}

@article{xu2025trajectorypredictionmeetslarge,
      title={{Trajectory Prediction Meets Large Language Models: A Survey}}, 
      author={Yi Xu and Ruining Yang and Yitian Zhang and Jianglin Lu and Mingyuan Zhang and Yizhou Wang and Lili Su and Yun Fu},
      year={2025},
      journal = { arXiv:2506.03408}

  }

@article{10.1145/2501654.2501656,
author = {Parent, Christine and Spaccapietra, Stefano and Renso, Chiara and Andrienko, Gennady and Andrienko, Natalia and Bogorny, Vania and Damiani, Maria Luisa and Gkoulalas-Divanis, Aris and Macedo, Jose and Pelekis, Nikos and Theodoridis, Yannis and Yan, Zhixian},
title = {{Semantic Trajectories Modeling and Analysis}},
year = {2013},
  volume = {45},
number = {4},
journal = {ACM Computing Survey},
}

@article{10.1145/2483669.2483682,
author = {Yan, Zhixian and Chakraborty, Dipanjan and Parent, Christine and Spaccapietra, Stefano and Aberer, Karl},
title = {{Semantic Trajectories: Mobility Data Computation and Annotation}},
year = {2013},
volume = {4},
number = {3},
journal = {ACM Transactions on Intelligent Systems and Technology}
}

@article{10.1145/3440207,
author = {Wang, Sheng and Bao, Zhifeng and Culpepper, J. Shane and Cong, Gao},
title = {{A Survey on Trajectory Data Management, Analytics, and Learning}},
year = {2021},
 volume = {54},
number = {2},
journal = {ACM Computing Survey}
}

@inproceedings{habitedbt,
  title={{Data-Driven Trajectory Imputation for Vessel Mobility Analysis}},
  author={Spiliopoulos, Giannis and Troupiotis-Kapeliaris, Alexandros and Patroumpas, Kostas and Liapis, Nikolaos and Skoutas, Dimitrios and Zissis, Dimitris and Bikakis, Nikos},
  booktitle=EDBT,
  year={2026}
}

@article{wang2026towards,
  author  = {Wang, Shen and Wu, Kejun and Qiao, Renjie and Wang, Runtian and Cai, Chengtao},
  title   = {{Towards Maritime Safety with Industrial Multimodal Models: Compliance Reasoning from Heterogeneous Maritime Data}},
  journal = {Applied Soft Computing},
  year    = {2026}
}

@article{yang2024harnessing,
  title={Harnessing the power of Machine learning for AIS Data-Driven maritime Research: A comprehensive review},
  author={Yang, Ying and Liu, Yang and Li, Guorong and Zhang, Zekun and Liu, Yanbin},
  journal={Transportation research part E: logistics and transportation review},
  volume={183},
   year={2024}
 }

@article{liu2026vista,
  title={{VISTA}: Knowledge-Driven Interpretable Vessel Trajectory Imputation via Large Language Models},
  author={Liu, Hengyu and Li, Tianyi and Wang, Haoyu and Torp, Kristian and Zhang, Tiancheng and Li, Yushuai and Jensen, Christian S},
  journal={arXiv:2601.06940},
  year={2026}
}

@article{GU2026101253,
title = {A survey on LLM-as-a-Judge},
journal = {The Innovation},
 year = {2026},
 author = {Jiawei Gu and Xuhui Jiang and Zhichao Shi and Hexiang Tan and Xuehao Zhai and Chengjin Xu and Wei Li and Yinghan Shen and Shengjie Ma and Honghao Liu and Saizhuo Wang and Kun Zhang and Zhouchi Lin and Bowen Zhang and Lionel Ni and Wen Gao and Yuanzhuo Wang and Jian Guo},
}

@article{bakdi2019ais,
  title={AIS-based multiple vessel collision and grounding risk identification based on adaptive safety domain},
  author={Bakdi, Azzeddine and Glad, Ingrid Kristine and Vanem, Erik and Engelhardtsen, {\O}ystein},
  journal={MDPI Journal of Marine Science and Engineering},
  year={2019},
}

\end{document}